%% file: main.tex
\input{preamble.tex}

\input{preamble_math.tex}

\input{preamble_acronyms.tex}

\usepackage{authblk}
\usepackage{enumitem} 
\usepackage{algorithmic}
\usepackage{tabularx}
\usepackage{booktabs}

\title{\textbf{Vendi Information Gain for Active Learning\\ and its Application to Ecology}}

\author[1, 2]{Quan Nguyen}
\author[1, 2, *]{Adji Bousso Dieng}
\affil[1]{Department of Computer Science, Princeton University}
\affil[2]{\href{https://vertaix.princeton.edu/}{Vertaix}}
    \affil[*]{Accepted at the AAAI Workshop on AI to Accelerate Science and Engineering (AI2ASE) 2026}

\begin{document}
\maketitle

\begin{abstract}
\noindent While monitoring biodiversity through camera traps has become an important endeavor for ecological research, identifying species in the captured image data remains a major bottleneck due to limited labeling resources. Active learning---a machine learning paradigm that selects the most informative data to label and train a predictive model---offers a promising solution, but typically focuses on uncertainty in the individual predictions without considering uncertainty across the entire dataset. We introduce a new active learning policy, Vendi information gain (VIG), that selects images based on their impact on dataset-wide prediction uncertainty, capturing both informativeness and diversity. We applied VIG to the Snapshot Serengeti dataset and compared it against common active learning methods.
VIG needs only 3\% of the available data to reach 75\% accuracy, a level that baselines require more than 10\% of the data to achieve. With 10\% of the data, VIG attains 88\% predictive accuracy, 12\% higher than the best of the baselines.
This improvement in performance is consistent across metrics and batch sizes, and we show that VIG also collects more diverse data in the feature space. VIG has broad applicability beyond ecology, and our results highlight its value for biodiversity monitoring in data-limited environments. \\

% \footnote{Package can be found at \url{https://github.com/vertaix/Vendi-Active-Learning}.}\\

\end{abstract}

\input{sec-introduction}
\input{sec-method}
\input{sec-experiments}
% \input{sec-related}
\input{sec-discussion}

% \subsection*{Acknowledgements}
% Adji Bousso Dieng is supported by the National Science Foundation, Office of Advanced Cyberinfrastructure (OAC): \#2118201 and DataX.

\section*{Dedication}
This paper is dedicated to \href{https://en.wikipedia.org/wiki/Wangar%C4%A9_Maathai}{Wangarĩ Muta Maathai}.

%\clearpage
\bibliographystyle{apa}
\bibliography{main}

\input{sec-appendix}

\end{document}

%% file: preamble.tex
\RequirePackage[l2tabu, orthodox]{nag}
\documentclass[11pt]{article}

\usepackage[T1]{fontenc}

\usepackage{soul} %% to use \st{} 

\usepackage[bitstream-charter]{mathdesign}
\usepackage{amsmath}
\usepackage[scaled=0.92]{PTSans}

\usepackage[
  paper  = letterpaper,
  left   = 1.65in,
  right  = 1.65in,
  top    = 1.0in,
  bottom = 1.0in,
  ]{geometry}

\usepackage[usenames,dvipsnames]{xcolor}
\definecolor{shadecolor}{gray}{0.9}

\usepackage[final,expansion=alltext]{microtype}
\usepackage[english]{babel}
\usepackage[parfill]{parskip}
\usepackage{afterpage}
\usepackage{framed}

{\endMakeFramed}

\usepackage{lineno}

\usepackage{ragged2e}

\newcounter{parcount}

\usepackage{graphicx}
\usepackage[labelfont=bf]{caption}
\usepackage{booktabs}
\usepackage{multirow}

\usepackage[algoruled]{algorithm2e}
\usepackage{listings}
\usepackage{fancyvrb}
\fvset{fontsize=\normalsize}

\usepackage{natbib}

\usepackage[colorlinks,linktoc=all]{hyperref}
\usepackage[all]{hypcap}
\hypersetup{citecolor=BurntOrange}
\hypersetup{linkcolor=black}
\hypersetup{urlcolor=MidnightBlue}

\usepackage[nameinlink]{cleveref}
\creflabelformat{equation}{#1#2#3}
\crefname{equation}{eq.}{eqs.}  
\Crefname{equation}{Eq.}{Eqs.}

\usepackage[acronym,nowarn]{glossaries}
\lstdefinestyle{mystyle}{
    commentstyle=\color{OliveGreen},
    keywordstyle=\color{BurntOrange},
    numberstyle=\tiny\color{black!60},
    stringstyle=\color{MidnightBlue},
    basicstyle=\ttfamily,
    breakatwhitespace=false,
    breaklines=true,
    captionpos=b,
    keepspaces=true,
    numbers=left,
    numbersep=5pt,
    showspaces=false,
    showstringspaces=false,
    showtabs=false,
    tabsize=2
}
\usepackage[colorinlistoftodos,
           prependcaption,
           textsize=small,
           backgroundcolor=yellow,
           linecolor=lightgray,
           bordercolor=lightgray]{todonotes}

\usepackage{amsthm}  % for proofs

\usepackage{bm}

\usepackage[colorinlistoftodos,
           prependcaption,
           textsize=small,
           backgroundcolor=yellow,
           linecolor=lightgray,
           bordercolor=lightgray]{todonotes}

\usepackage{graphicx}
\usepackage{caption}
\usepackage{subcaption}
\usepackage{wrapfig}

\usepackage{booktabs}
\usepackage{arydshln} % Dashed lines
\usepackage{multirow}

\usepackage{listings}
\usepackage{fancyvrb}
\fvset{fontsize=\normalsize}

\usepackage[colorlinks,linktoc=all]{hyperref}
\hypersetup{linkcolor=black}
\hypersetup{urlcolor=BurntOrange}

\usepackage[nameinlink]{cleveref}
\creflabelformat{equation}{#1#2#3}
\crefname{equation}{eq.}{eqs.}  
\Crefname{equation}{Eq.}{Eqs.}

\usepackage{listings}
\lstdefinestyle{alp_style}{
    commentstyle=\color{OliveGreen},
    numberstyle=\tiny\color{black!60},
    stringstyle=\color{BrickRed},
    basicstyle=\ttfamily\scriptsize,
    breakatwhitespace=false,
    breaklines=true,
    captionpos=b,
    keepspaces=true,
    numbers=none,
    numbersep=5pt,
    showspaces=false,
    showstringspaces=false,
    showtabs=false,
    tabsize=2
}

\usepackage{capt-of}

\usepackage{lipsum}

\theoremstyle{remark}
\newtheorem*{lemma*}{Lemma}

%% file: preamble_math.tex
\def\eqref#1{equation~\ref{#1}}
\def\1{\bm{1}}

\DeclareMathAlphabet{\mathsfit}{\encodingdefault}{\sfdefault}{m}{sl}
\SetMathAlphabet{\mathsfit}{bold}{\encodingdefault}{\sfdefault}{bx}{n}

%% file: preamble_acronyms.tex
\newacronym{ALI}{ali}{adversarially learned inference}
\newacronym{BIGAN}{bigan}{bidirectional generative adversarial network}
\newacronym{VI}{vi}{variational inference}
\newacronym{KL}{kl}{Kullback-Leibler}
\newacronym{ELBO}{elbo}{evidence lower bound}
\newacronym{MCMC}{mcmc}{Markov chain Monte Carlo}
\newacronym{HMC}{hmc}{Hamiltonian Monte Carlo}
\newacronym{RNN}{rnn}{recurrent neural network}
\newacronym{MLP}{mlp}{feed forward neural network}
\newacronym{GAN}{gan}{generative adversarial network}
\newacronym{DCGAN}{dcgan}{deep convolutional generative adversarial network}
\newacronym{PresGAN}{presgan}{prescribed generative adversarial network}
\newacronym{DGM}{dgm}{deep generative model}
\newacronym{PGAN}{pgan}{prescribed generative adversarial network}
\newacronym{VEEGAN}{veegan}{vee {GAN}}
\newacronym{PACGAN}{pacgan}{packed {GAN}}
\newacronym{STYLEGAN}{stylegan}{Style {GAN}}
\newacronym{FID}{fid}{{F}r\'{e}chet {I}nception distance}
\newacronym{IS}{is}{{I}nception score}
\newacronym{ML}{ml}{machine learning}
\newacronym{VS}{vs}{vendi score}
\newacronym{NLP}{nlp}{natural language processing}
\newacronym{IntDiv}{intdiv}{{I}nternal {D}iversity}
\newacronym{BLEU}{bleu}{BLEU}
\newacronym{PAIRWISE-BLEU}{pairwise-bleu}{PAIRWISE-BLEU}
\newacronym{D-LEX-SIM}{d-lex-sim}{D-LEX-SIM}
\newacronym{GILBO}{gilbo}{GILBO}
\newacronym{NOM}{nom}{number of modes}
\newacronym{MOSES}{moses}{MOSES}
\newacronym{HMM}{hmm}{HMM}
\newacronym{AAE}{aae}{AAE}
\newacronym{VAE}{vae}{VAE}
\newacronym{JTN}{jtn}{JTN}
\newacronym{Char-RNN}{char-rnn}{Char-RNN}
\newacronym{SMILES}{smiles}{SMILES}
\newacronym{MNIST}{mnist}{MNIST}
\newacronym{MultiNLI}{multinli}{MultiNLI}
\newacronym{StackedMNIST}{stackedmnist}{StackedMNIST}
\newacronym{NLI}{nli}{NLI}
\newacronym{VDVAE}{vdvae}{VDVAE}
\newacronym{LSUN}{lsun}{LSUN}
\newacronym{CIFAR}{cifar}{CIFAR}
\newacronym{ADM}{adm}{ADM}
\newacronym{RQ-VT}{rq-vt}{RQ-VT}
\newacronym{IDDPM}{iddpm}{IDDPM}
\newacronym{DBS}{dbs}{diverse beam search}
\newacronym{MS COCO}{ms coco}{MS COCO}
\newacronym{CelebA}{celeba}{CELEBA}
\newacronym{CIFAR-100}{cifar-100}{CIFAR-100}
\newacronym{DRM}{drm}{diversity risk minimization}
\newacronym{ERM}{erm}{empirical risk minimization}
\newacronym{VRM}{vrm}{vicinal risk minimization}
\newacronym{MIXUP}{mixup}{MIXUP}
\newacronym{Vendi-VI}{vendi-vi}{Vendi Variational Inference}
\newacronym{Vendi-VAE}{vendi-vae}{Vendi Variational Auto-Encoder}
\newacronym{VeT}{VeT}{Vendi Transform}

%% file: sec-introduction.tex
%!TEX root = main.tex

\section{Introduction}
\glsresetall

The ability to monitor biodiversity at scale is critical for understanding ecosystem health and informing conservation efforts.
Camera traps---remotely activated cameras triggered by motion or heat---have become a key tool for ecological data collection, enabling large-scale, non-invasive monitoring of wildlife in their natural habitats~\citep{trolliet2014use,delisle2021next,tuia2022perspectives}.
These devices generate vast volumes of image data, often spanning multiple times of day and geographies.
However, the subsequent task of identifying and labeling the species in these images remains a significant bottleneck.
Manual annotation is labor-intensive, costly, time-consuming, and may require expert knowledge, especially when dealing with rare species or poor image quality.

Recent advances in machine learning, specifically deep learning for image classification, offer a promising direction for automating species identification~\citep{norouzzadeh2018automatically,beery2018recognition}.
Yet the performance of these models crucially depends on the availability of large amounts of high-quality labeled training data.
In many ecological applications, however, labels are scarce and labeling is costly. % are scarce and the labeling budgets are constrained.
% Further, many datasets exhibit long-tailed distributions, with common species dominating the data and rare ones underrepresented.
These challenges motivate the need for intelligent sampling strategies that maximize model performance while minimizing labeling effort.

\begin{figure}[!t]
\centering
\includegraphics[width=\linewidth]{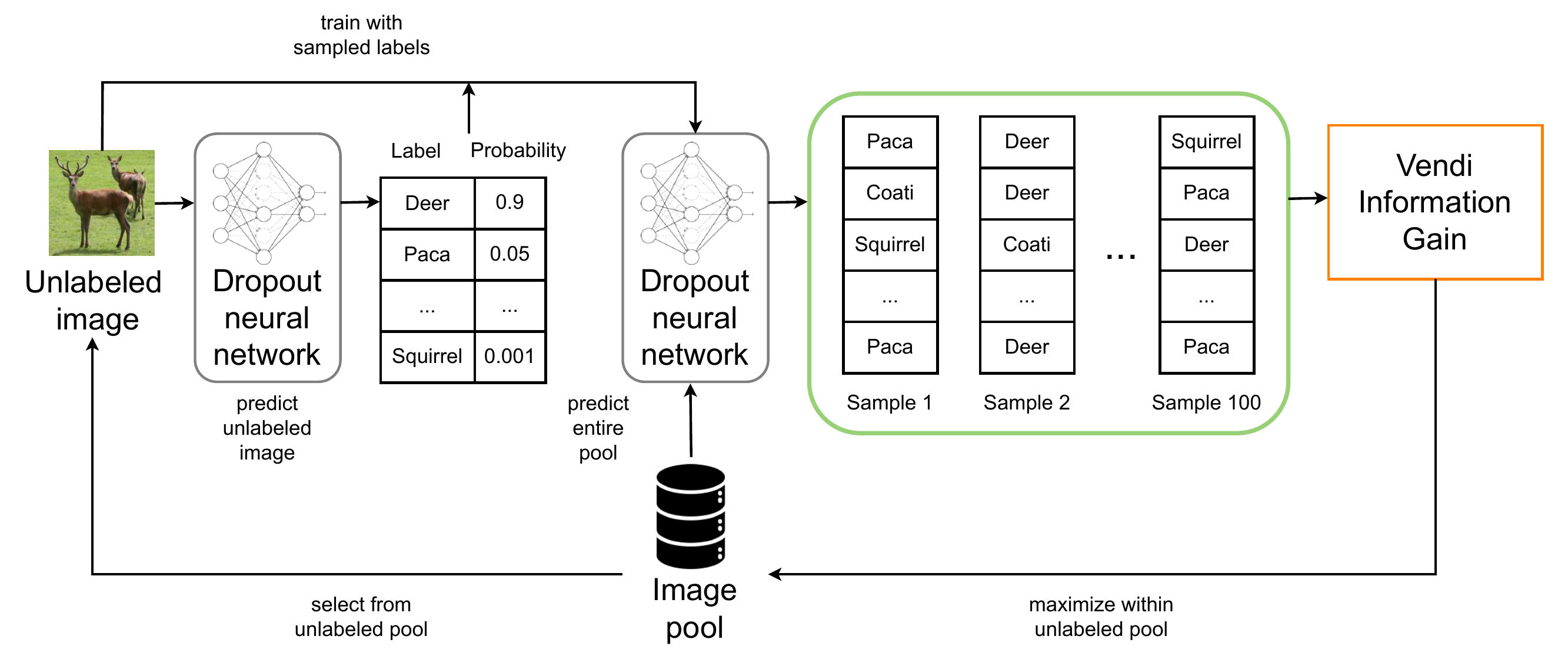}
\caption{
Overview of Vendi information gain (VIG) for active learning.
We use a trained dropout neural network to sample labels for a candidate datapoint.
The neural network is then retrained on this fantasized data to sample labels of the entire pool.
Uncertainty in these predictions is captured by VIG, and we select the candidate that yields the highest information gain (i.e., lowest uncertainty) in the predictions to label.
The result is then added to the training dataset, and the process repeats until the labeling budget is exhausted.
}
\label{fig:diagram}
\end{figure}

Active learning~\citep{settles2009active,bothmann2023automated} offers a principled solution to this problem.
By iteratively selecting the most informative examples to label, active learning algorithms can achieve high accuracy with fewer labeled instances than na\"ive approaches.
Existing active learning solutions reason about the level of informativeness of candidates for labeling on an individual basis---targeting datapoints that the model is most uncertain about---without accounting for the effects of those datapoints post-labeling.
This perspective neglects the overall structure of the entire image pool.
%In ecology, where rare species and diverse image conditions are crucial, being able to account for all datapoints could prove beneficial.

% In this work, we propose a novel active learning policy based on the recent information gain criterion called Vendi Information Gain (VIG), designed to guide labeling efforts in ecological monitoring projects.
% Our approach reasons about the amount of information about the label of \emph{all} images within the pool gained to identify high-value images for annotation.
% Figure~\ref{fig:diagram} shows the schematics of VIG.
% We evaluate it using a challenging image classification task using images from the Snapshot Serengeti dataset~\citep{swanson2015snapshot}, which provides camera trap images of various species in a savanna.
% We demonstrate that VIG finds diverse data that effectively informs the model, and as a result consistently outperforms baselines in terms of label efficiency and predictive accuracy.
In this work, we propose Vendi information gain (VIG), a novel active learning policy designed to optimize the global informativeness of the training data.
VIG builds on recent advances in information-theoretic metrics and quantifies the reduction in predictive uncertainty across the entire image pool when a candidate image is labeled.
This approach selects datapoints not only because they yield high uncertainty, but because they are likely to inform the model's predictions across the board.
Figure~\ref{fig:diagram} shows the schematics of VIG consisting of the following steps.
First, we sample candidate labels for each unlabeled image using a dropout neural network predictor.
We then retrain the model on these fantasized labels and sample predictions for the entire unlabeled pool.
These sampled predictions quantify the reduction in Vendi entropy~\citep{friedman2023vendi,nguyen2025vendi} across the unknown labels, which guides the search for the candidate with the highest information gain.
This process repeats iteratively, expanding the labeled set until the labeling budget is exhausted.
The use of a dropout neural network for active learning is described in Section~\ref{sec:dropout}, and Section~\ref{sec:vig-al} includes the computational details of VIG.
% [EXPLAIN HERE THE OVERALL PIPELINE AND POINT TO THE APPROPRIATE SUBSECTION IN THE METHODS SECTION FOR THE DETAILS]

Applied to the Snapshot Serengeti dataset~\citep{swanson2015snapshot}---a benchmark for camera trap classification---VIG consistently outperforms standard active learning baselines in terms of label efficiency and predictive accuracy.
We show that VIG collects more diverse data in feature space, leading to better generalization with fewer labels.
Our results suggest that VIG can serve as a general-purpose method for data-efficient ecological monitoring. %, especially when data diversity is crucial.

%%% Local Variables:
%%% mode: latex
%%% TeX-master: "main"
%%% End:

%% file: sec-method.tex
%!TEX root = main.tex

\section{Method}
\label{sec:method}

We first discuss the active learning framework and the use of a dropout neural network as the predictor for this task.
We then provide background on VIG as a metric of information gain and present its adoption to active learning.

\subsection{Active Learning Policies}

Active learning targets the common setting in machine learning  where labeling data is costly (in terms of time, money, or some safety-critical conditions).
The goal is to design an active learning policy that selects a small amount of data to label, so that the predictive model trained on the labeled data achieves good generalization performance.
In our setting, we have access to a large database of unlabeled images $\mathcal{X} = \{ x_i \}_{i=1}^n$, where each $x_i$ denotes a particular datapoint (image) within the database.
These images are classified into a predetermined number of classes $[C] = \{ 1, 2, \ldots, C\}$, and the unknown label $y_i$ of datapoint $x_i$ denotes the membership of that point.
Active learning proceeds in an iterative manner where at each step, the active learning policy selects a batch of images to label, adding them to the training data.
The process repeats such that we accumulate a training set of increasing size until our labeling budget is depleted.

The main focus of active learning is the design of the policy that selects which data to label.
Increasing information (or decreasing uncertainty) in the knowledge of the trained model serves as a popular heuristic for this task.
Formally, assume that we have a probabilistic model that produces the posterior probability that a point $x \in \mathcal{X}$ belongs to class $c \in [C]$, denoted as $p(y = c \mid x)$.
(We omit the dependence on the labeled data $\mathcal{D}$ that the model is trained on for conciseness.)
Many active learning policies seek to minimize model uncertainty, quantified by various statistics from the predictive distribution $p(y \mid x)$.
For instance, the \emph{Max entropy} policy finds the data that have the highest predictive entropy $H$~\citep{shannon1948mathematical} to quantify uncertainty in the predictions:
\begin{equation}
H(y \mid x) = -\sum_{c \in [C]} p(y=c \mid x) \log p(y=c \mid x).
\end{equation}
Other policies target alternative ways to quantify predictive uncertainty.
This includes the \emph{Mean STD} policy targeting the average standard deviation in the predictions:
\begin{equation}
\sigma(x) = \frac{1}{C} \sum_{c \in [C]} \sqrt{\mathbb{E} \left[ p(y = c \mid x)^2 \right] - \mathbb{E} \left[ p(y = c \mid x) \right]^2},
\end{equation}
which corresponds to the standard deviation statistic in the regression setting, but has been recently adopted in classification as well~\citep{kampffmeyer2016semantic,kendall2017bayesian}.
Another popular active learning policy, BALD, maximizes the amount of information gained about the predictive model's parameters $\boldsymbol{\omega}$, which is equivalent to maximizing the mutual information $I$ between predictions and model posterior~\citep{houlsby2011bayesian}:
\begin{equation}
I(\boldsymbol{\omega}, y \mid x) = H(\boldsymbol{\omega}) - \mathbb{E}_{p(y \mid x)} \left[ H(\boldsymbol{\omega} \mid x, y) \right].
\end{equation}
Finally, \citet{kirsch2019batchbald} proposed BatchBALD that extends BALD to account for interactions between datapoints within a batch.
We use these active learning policies as baselines to compare VIG against.

\subsection{Dropout Neural Networks}
\label{sec:dropout}

The previously described active learning policies depend on a probabilistic model producing predictions of the form $p(y = c \mid x)$, and as such have been limited to kernel-based methods such as Gaussian processes~\citep{li2013adaptive}.
In the context of image classification, these methods require a kernel to operate on images, which do not scale well to high-dimensional data or capture spatial information within the input images.
On the other hand, convolutional neural networks~\citep{rumelhart1985learning,lecun1989backpropagation} have proven to be effective at learning from images and achieved human-level performance at image recognition.
However, neural networks do not inherently produce probabilistic predictions with calibrated uncertainty quantification.

% Initially developed as a technique to regularize neural networks, dropout~\citep{hinton2012improving,srivastava2014dropout} has been used to approximate Bayesian inference with these neural networks.
% In dropout, random nodes in the hidden layers of a neural network are disabled at each forward pass during training, which 
% \citet{gal2016dropout} showed that 
Initially developed to regularize neural networks, dropout~\citep{hinton2012improving,srivastava2014dropout} dictates that random nodes in the hidden layers of a neural network are disabled at each forward pass during training.
\citet{gal2016dropout} further showed that using dropout during inference produces Monte Carlo samples from the predictive distribution of the corresponding Bayesian neural network trained with variational inference, naming the technique \emph{MC dropout}.
Finally, \citet{gal2017deep} used MC dropout as the predictive model to perform active learning on high-dimensional image data, showing that combined with MC dropout, the policies previously described outperform kernel-based active learning methods as well as their counterparts that use the predictions of a non-dropout neural network.
We use this neural network model with MC dropout as the probabilistic classifier in our experiments.

\subsection{Vendi Information Gain}

VIG was based on the Vendi Score (VS), a flexible diversity metric.
First proposed by \citet{friedman2023vendi} and later extended by \citet{pasarkar2024cousins}, the VS operates on a set of datapoints $D = \{ \theta_i \}_{i = 1}^n$ sampled from some domain $\Theta$.
To realize the VS, we first require a positive semidefinite kernel function $k: \Theta \times \Theta \rightarrow \mathbb{R}$, where $k(\theta, \theta) = \nolinebreak 1, \forall \theta \in \Theta$.
We then compute the kernel matrix $K \in \mathbb{R}^{n \times n}$, where each entry $K_{i, j} = k(\theta_i, \theta_j)$.
Finally, we define the VS as:
\begin{equation}
\label{eq:vs_q}
\mathrm{VS}_q (D; k) = \exp \left( \frac{1}{1 - q} \, \log \biggl( \sum_{i = 1}^n (\overline{\lambda}_i)^q \biggr) \right).
\end{equation}
where $\overline{\lambda}_1, \overline{\lambda}_2, \ldots, \overline{\lambda}_n$ are the eigenvalues of $K$, normalized so that they sum to $1$, and the order $q \geq 0$ is a hyperparameter.
The VS has since been extended and applied to various domains, including evaluating generative models~\citep{hall2024towards,senthilkumar2024beyond,jalali2024conditional}, molecular simulations~\citep{pasarkar2023vendi}, Bayesian optimization~\citep{liu2024diversity} and active search~\citep{nguyen2024quality}, sequence generative models~\citep{rezaei2025alpha}, RAG approaches for LLMs~\citep{rezaei2025vendi}, analysis of large-scale data~\citet{pasarkar2025vendiscope}, and reinforcement learning~\citep{lintunen2025vendirl}.

In particular, \citet{nguyen2025vendi} introduced VIG as a metric of information gain, defining it as the difference in the Vendi entropy $H_V$ of a random variable $\theta$ before and after conditioning on another variable $y$:
\begin{equation}
\label{eq:vig}
\mathrm{VIG}(\theta, y; q) = H_V(D; q) - \mathbb{E}_y[H_V(D_y; q)],
\end{equation}
where $D = \{ \theta_i \}_{i = 1}^n$ is a set of samples of $\theta$, and $D_y = \{ \theta_i \mid y \}_{i = 1}^n$ is the corresponding set of samples conditioned on a particular value of $y$.
Here, the Vendi entropy is the logarithm or the VS, or the R\'enyi entropy of the normalized eigenvalues of the kernel matrix computed from a set of samples:
\begin{equation}
\label{eq:von_neumann_entropy}
H_V(D; q) = \frac{1}{1 - q} \, \log \biggl( \sum_{i = 1}^n (\overline{\lambda}_i)^q \biggr).
\end{equation}

\citet{nguyen2025vendi} demonstrated many of VIG's advantages over mutual information, the default measure of information gain in the scientific literature~\citep{shannon1948mathematical,cover1999elements}. Namely, VIG works well with only samples of the random variable of interest and offers a more principled quantification of information gain that accounts for sample similarity.
The authors showcased VIG's superior performance in a wide range of tasks, including experimental design problems and level-set estimation.

\subsection{Vendi Information Gain for Active Learning}
\label{sec:vig-al}

We adopt the VIG criterion for active learning, proposing a policy that minimizes the Vendi entropy of the posterior predictions across the entire database of images, conditioned on a candidate datapoint.
Formally, denote $\boldsymbol{\theta}$ as the vector that concatenates the unknown labels of the images within the database, the VIG policy finds the datapoint $x$ that minimizes the posterior Vendi entropy in $\boldsymbol{\theta}$:
\begin{equation}
\mathrm{VIG}(\boldsymbol{\theta}, x) = H_V(D) - \mathbb{E}_{y \mid x} \left[ H_V(D_y \mid x) \right],
\end{equation}
where $D$ is a set of samples of the label vector $\boldsymbol{\theta}$, and $D_y$ is the corresponding set of samples conditioned on a particular label $y$ of image $x$.
These samples can be generated using the MC dropout neural network when predicting on the images in the database.

The computation of Vendi entropy requires a kernel that compares two given sample label vectors $\boldsymbol{\theta}_1$ and $\boldsymbol{\theta}_2$.
We compute the Hamming distance $d_H$ between these vectors and subtract the normalized distance from $1$ to produce a similarity measure:
\begin{equation}
k(\boldsymbol{\theta}_1, \boldsymbol{\theta}_2) = 1 - \frac{d_H(\boldsymbol{\theta}_1, \boldsymbol{\theta}_2)}{N},
\end{equation}
where $N$ is the length of the label vectors.
Note that this is not the same kernels in the kernel-based active learning policies, which seek to operate on the images themselves.

This choice of kernel is natural, as two labels are similar to each other only if they belong to the same class.
When there is only one datapoint in the pool, the Vendi entropy induced by this kernel coincides with the Shannon entropy of the datapoint's class distribution---a reassuring feature.

Overall, while traditional active learning policies target individual predictive uncertainty measures, VIG selects datapoints expected to reduce uncertainty in predictions over the entire unlabeled pool, accounting for global informativeness.
To compute the VIG score for a candidate image $x$, we sample possible labels $y$, retrain the model with the labeled $x$, then sample predictions $\boldsymbol{\theta}$ for the full pool.
The candidate with the highest VIG score is selected.

%%% Local Variables:
%%% mode: latex
%%% TeX-master: "main"
%%% End:

%% file: sec-experiments.tex
%!TEX root = main.tex

\section{Experiments}
\glsresetall

\begin{figure}[!t]
\centering
\includegraphics[width=\linewidth]{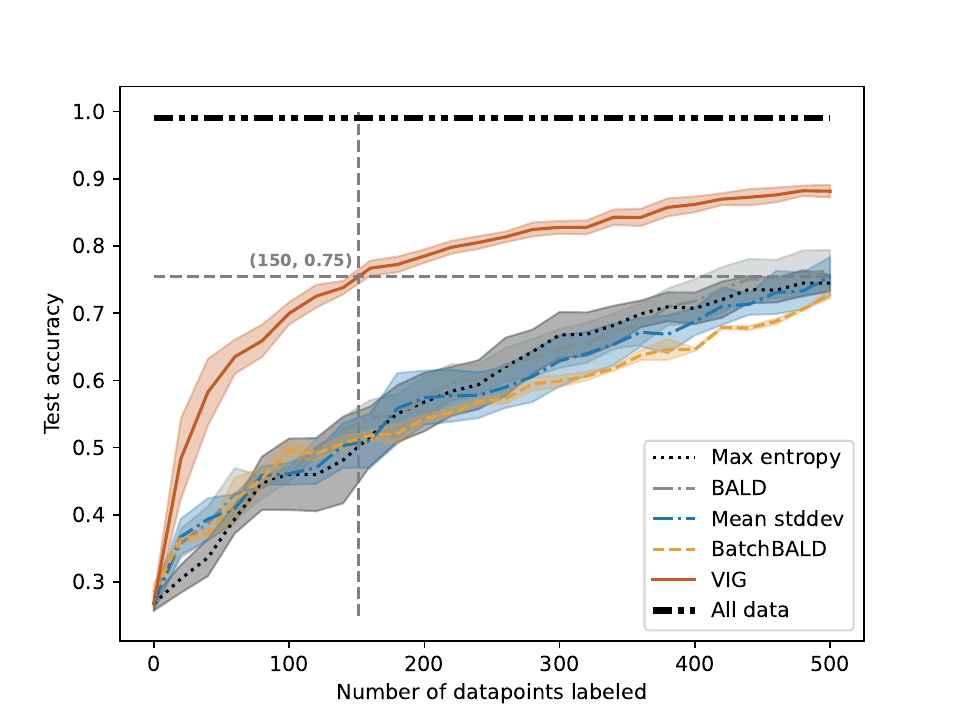}
\caption{
Average test accuracy ($\pm 1$ standard error) by various active learning policies.
VIG obtains a large gain right from the start and maintains its lead throughout the active learning loop.
It takes VIG only 150 datapoints to achieve the accuracy of 75\% that other methods need 500 points to achieve.
Meanwhile, at 500 points, VIG achieves close to 90\% accuracy.
In comparison, training on all available training data (5000+ images) yields an accuracy of 99\%.
}
\label{fig:classification}
\end{figure}

We benchmark our method VIG against existing baselines in active learning described in Section~\ref{sec:method}.
At each iteration of the active learning loop, each policy obtains a batch of 20 images to label, and the process repeats until 500 images are collected.

% Our goal is to train a model to accurately predict the species of the animal appearing in a camera trap image.
% The difficulty of accurately labeling an image, which requires human expertise, motivates the active learning approach.
Figure~\ref{fig:classification} shows the accuracy on a hold-out test set of the model trained on data collected by various active learning policies, averaged across repeated, as a function of the number of datapoints labeled.
VIG significantly outperforms the baselines, achieving a higher accuracy with fewer labels.
After obtaining 500 labeled datapoints, VIG yields a test accuracy close to 90\%, while other policies reach 75\%.
To achieve the same performance, VIG needs only 150 labels.
In comparison, assuming unlimited labeling resources, the model trained on all available training data (5585 images) yields a test accuracy of 99\%.

\begin{figure}[!t]
\centering
\includegraphics[width=\linewidth]{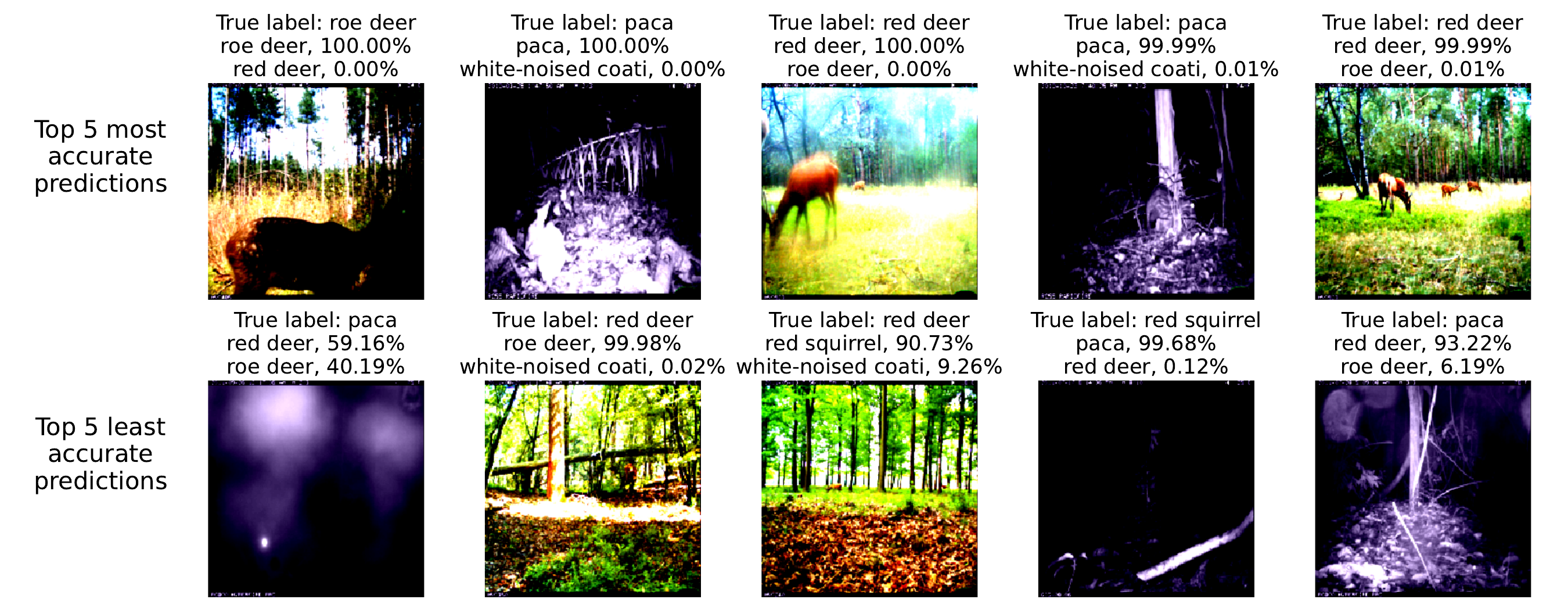}
\caption{
The 5 most (top row) and 5 least (bottom row) accurate predictions by the model trained with data collected by VIG.
In the bottom row, the model understandably makes mistakes on instances where the animal is barely visible.
}
\label{fig:show_preds}
\end{figure}

To inspect the learned behavior of VIG's model, Figure~\ref{fig:show_preds} visualizes the top five most and least accurate predictions on the test set by VIG.
On the top row that the trained model reassuringly makes accurate predictions with high confidence on instances where the animal is visibly in the middle of the camera trap images.
On the bottom row, the model understandably makes mistakes on instances with low visibility, including those taken in the dark---situations even humans find challenging.

\begin{table}[t]
\centering
\caption{
Average test statistics by various active learning policies at 500 labeled datapoints.
(Recall is omitted as it coincides with accuracy by definition.)
VIG consistently outperforms the baselines across the different metrics.
}
\label{tab:result}
\begin{tabular}{cccccc}
\toprule
& Max entropy & BALD & Mean stddev & BatchBALD & VIG \\
\midrule
% Accuracy $\uparrow$ & 0.745 & 0.763 & 0.755 & 0.728 & \textbf{0.882} \\
Precision $\uparrow$ & 0.780 & 0.799 & 0.780 & 0.764 & \textbf{0.888} \\
F1 score $\uparrow$ & 0.755 & 0.775 & 0.765 & 0.738 & \textbf{0.883} \\
Cross-entropy loss $\downarrow$ & 0.705 & 0.635 & 0.625 & 0.707 & \textbf{0.402} \\
\bottomrule
\end{tabular}
\end{table}

Table~\ref{tab:result} lists other metrics of classification performance than accuracy in Figure~\ref{fig:classification}.
This includes the cross-entropy loss, which accounts for the model's predictive confidence, rewarding confident correct predictions and punishing confident incorrect ones.
Overall, VIG consistently achieves the best performance across the metrics.

\begin{figure}
\centering
\includegraphics[width=\linewidth]{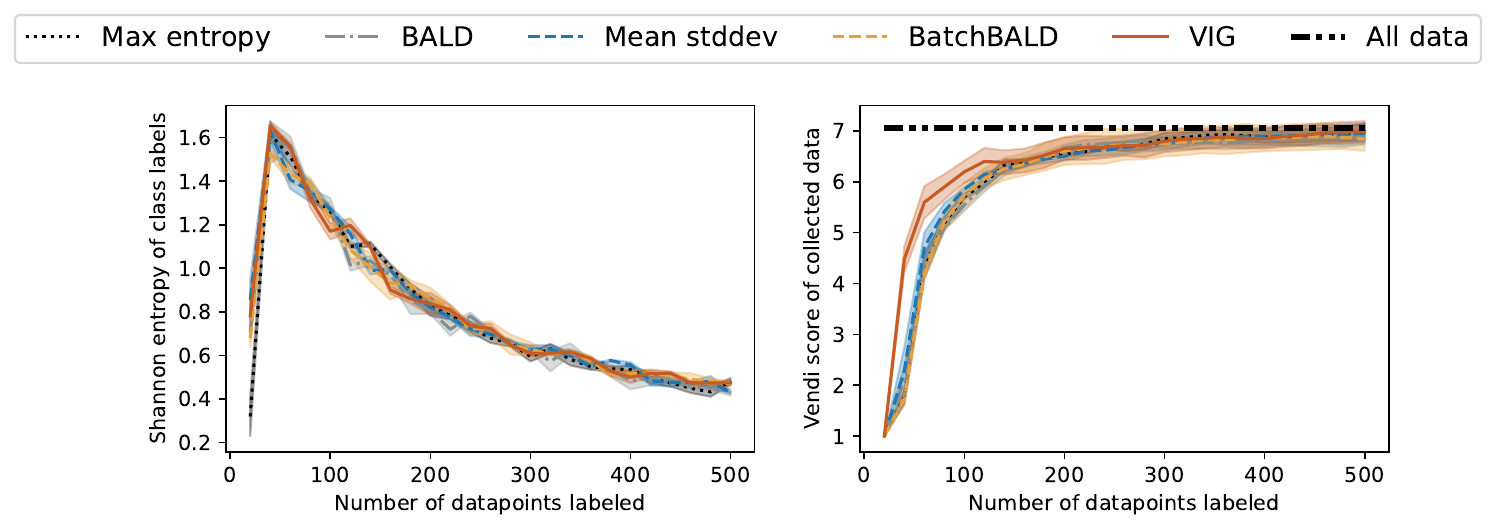}
\caption{
Diversity of the collected data by various active learning policies.
\textbf{Left}: The Shannon entropy of the class distribution of the collected data.
Here, all methods are comparable.
\textbf{Right}: The Vendi score of the collected data using the embedding in the second-to-last layer of the neural network classifier trained on all available data.
VIG selects more diverse data right from the beginning.
}
\label{fig:diversity}
\end{figure}

To understand what drives VIG's performance, we inspect the diversity of the data collected by each method in Figure~\ref{fig:diversity}.
The left panel shows diversity in the labels of the collected data, measured by the Shannon entropy of the class distribution.
Here, all policies are comparable.
In the right panel we show diversity in the features of the images, quantified by the Vendi score (VS)~\citep{friedman2023vendi} of the labeled images.
The VS is a flexible diversity metric whose output has the natural interpretation of the effective number of unique elements in a set.
The VS requires a kernel function to compute the similarity of two given datapoints.
Following previous works~\citep{friedman2023vendi,pasarkar2024cousins,askari2024improving}, we choose the cosine kernel operating on the image embedding.
To have a consistent embedding across different active learning policies, we train a neural network classifier on all available data and use the features in the second-to-last layer.
Right from the start of the active learning loop, VIG collects more diverse data (feature-wise), a behavior previous works have demonstrated to be beneficial for active learning~\citep{yang2015multi,du2015exploring,buchert2022exploiting}.

\begin{figure}
\centering
\includegraphics[width=\linewidth]{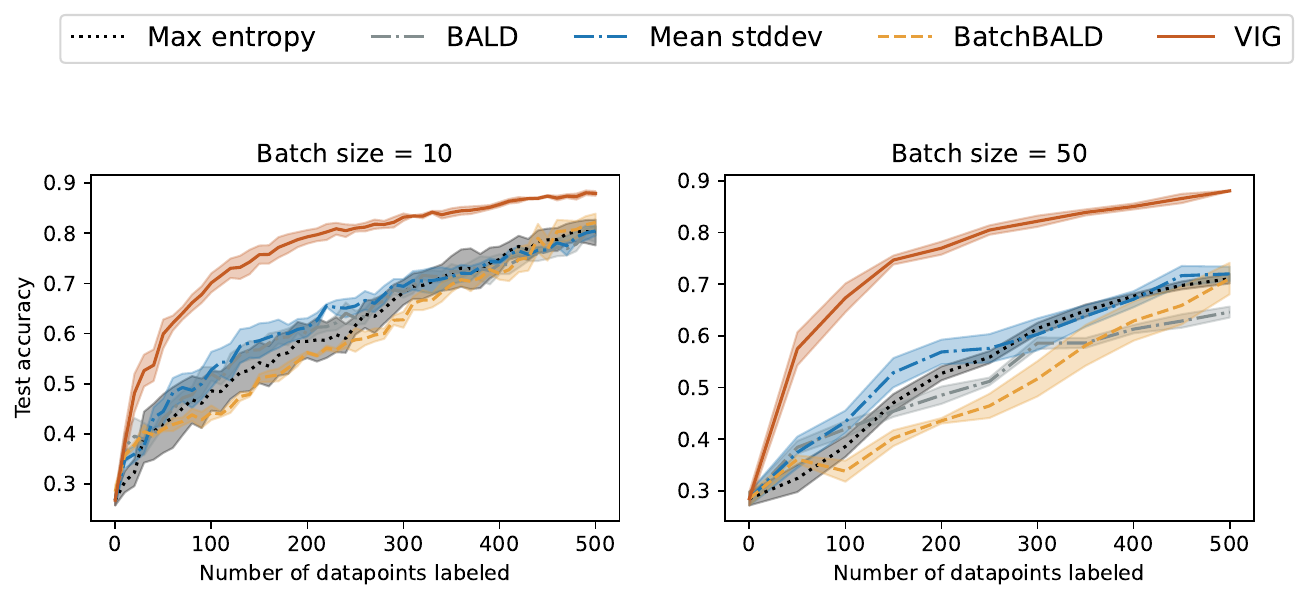}
\caption{
Test accuracy by various active learning policies under different batch sizes.
VIG's superior performance stays consistent in both low- and high-throughput settings, underscoring its robustness to selection frequency.
}
\label{fig:batch_size}
\end{figure}

Figure~\ref{fig:classification} shows the performance of the active learning policies when the batch size (the number of images selected to be labeled at each step of the learning loop) is set to 20.
We repeat these experiments while varying this batch size to investigate the effect of this parameter.
The left panel of Figure~\ref{fig:batch_size} shows the same results under batch size 10, representing a low-throughput setting, while the right panel gives batch size 50 (a high-throughput setting).
We see the reasonable trend that policies tend to perform better when the batch size is small, as they get more frequent feedback from the labels and thus can be more adaptive in their selections.
Further, VIG stays competitive across the different batch sizes, illustrating the benefits of our method.

These results collectively show that VIG's reasoning allows it to extract more information from fewer labels, making it suitable for ecological settings with limited annotation budgets.

%%% Local Variables:
%%% mode: latex
%%% TeX-master: "main"
%%% End:

%% file: sec-discussion.tex
%!TEX root = main.tex
\glsresetall

\section{Limitations}

VIG requires evaluating fantasized scenarios for each candidate image.
For every possible label, the method (1) retrains the model on the augmented training set, (2) generates $n$ posterior label samples over the unlabeled pool of size $N$ via $T$ forward passes of MC dropout that scales like $O(TN)$, and (3) computes Vendi entropy, which includes an eigen-decomposition of an $n \times n$ similarity matrix that scales like $O(n^3)$.
The overall complexity scales like $O\big(m(R + TN + n^3) \big)$, where $m$ is the number of fantasized labels we use when considering each candidate image, and $R$ is the cost of retraining the model on each fantasized label.
In our experiments, we fantasize using the top $m = 3$ most-likely labels and set $n = 100$.

To make the method more efficient, we employ early stopping for the retraining step, terminating the training process early if the training loss converges.
Our justification is that during VIG's computation, as we add one single sampled label to the training set, the model trained at the previous step is already close to an optimum.
Further, due to the difficulty in obtaining labels in active learning settings, the size of the training set is often limited, which allows for faster training.
In our experiments, VIG takes about 4 seconds per evaluation---an acceptable speed given the boost in performance from the method.

Under extremely large datasets, we can sub-sample the unlabeled pool and conduct the current iteration's search within the sampled subset.
This sub-sampling technique was studied in \citet{mirzasoleiman2015lazier} and used in other active learning settings~\citep{nguyen2021nonmyopic,nguyen2023nonmyopic}.

\section{Discussion}
\label{sec:discussion}

We study a new active learning policy, Vendi information gain (VIG), and demonstrate its effectiveness in image-based biodiversity monitoring.
By selecting images that maximize information gain over the entire unlabeled pool, VIG prioritizes examples that not only have high uncertainty but are also informative and diverse.
With camera trap data from the Snapshot Serengeti dataset, VIG achieves substantial gains in label efficiency and predictive performance compared to established baselines.

Though we focus on species classification from camera trap images, VIG is general-purpose and model-agnostic.
% That said, the method can be readily applied to domains beyond ecology---it simply requires a probabilistic model to sample from, a role that a dropout neural network like ours or a Gaussian process can play.
The method only requires a probabilistic predictor capable of generating samples, such as a dropout neural network like ours or a Gaussian process.
This makes VIG generalizable to a broad range of tasks beyond ecological applications.

VIG's superior performance highlights the value of using structured diversity to quantify uncertainty---an approach that aligns well with the complexity and richness of ecological data.
Future work may explore its application in regression tasks such as estimating the abundance of species, or integration with crowd-sourced labeling platforms to elicit expert labeling effort when it is most needed.

%%% Local Variables:
%%% mode: latex
%%% TeX-master: "main"
%%% End:

%% file: sec-appendix.tex
%\section{Appendix}

\clearpage
\appendix

\section{Additional Experiment Results}
% \label{app:proofs}

\begin{figure}
\centering
\includegraphics[width=\linewidth]{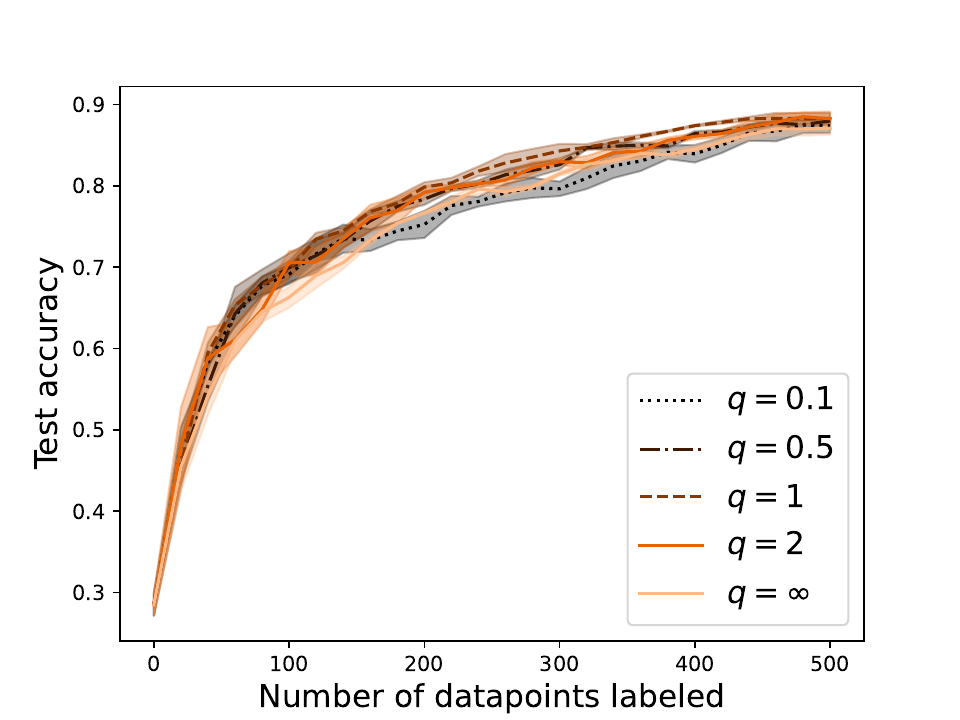}
\caption{
Average test accuracy and one standard error by VIG of different orders $q$.
VIG's performance is robust against the value of $q$.
}
\label{fig:order}
\end{figure}

We now present the result of an ablation study where we investigate the effect of the hyperparameter $q$ in the formulation of VIG.
\citet{pasarkar2024cousins} showed that the order $q$ controls the sensitivity of the Vendi score (and thus the Vendi entropy and VIG) to rarity: low values of $q$ lead to more sensitivity to rare features, while high values of $q$ prioritize common features of the samples.
By setting this hyperparameter, we can induce a family of VIG policies with different levels of sensitivity to rare samples.
Figure~\ref{fig:order} shows the results of VIG across a wide range of values for $q$, where test performance is comparable across the VIG policies.
This shows that the performance improvement from existing active learning baselines we obtain is mainly due to VIG's information gain-based reasoning, which is robust against the order $q$ when computing Vendi entropy.

\section{Data}

We use the latest iteration of the Snapshot Serengeti dataset~\citep{swanson2015snapshot} and extract the top five species, making a dataset with the following class breakdown:
\begin{itemize}
\item Paca: 1196 images
\item Red deer: 2830 images
\item Red squirrel: 639 images
\item Roe deer: 1271 images
\item White-nosed coati: 1295 images
\end{itemize}

This makes up a 7231-image dataset.
In each experiment, we pick a randomly selected 20\% of the data as the test set to evaluate the trained models; the other 80\% acts as the image pool for active learning.

%%% Local Variables:
%%% mode: latex
%%% TeX-master: "main"
%%% End: